\documentclass{article}

     \PassOptionsToPackage{square,sort,comma,numbers}{natbib}

     \usepackage[final]{neurips_wrl2020}

\usepackage[utf8]{inputenc} 
\usepackage[T1]{fontenc}    
\usepackage{hyperref}       
\usepackage{url}            
\usepackage{booktabs}       
\usepackage{amsfonts}       
\usepackage{nicefrac}       
\usepackage{microtype}      

\usepackage{amsmath}
\usepackage{graphicx}
\usepackage{subfigure}

\usepackage{enumerate}
\usepackage{amsmath}
\usepackage{mathtools}
\usepackage{stackrel}
\usepackage{listings}
\usepackage{flushend}
\usepackage{amsfonts}
\usepackage{array}
\usepackage{forest}
\usepackage{array}
\usepackage{forest}
\usepackage{tikz}
\usepackage{algorithm2e}
\usepackage{url}
\usepackage{tkz-graph}  
\usepackage{blindtext}
\usepackage{gensymb}
\usepackage{comment}
\usepackage{lipsum}
\usepackage{mathtools}
\usepackage{cuted}
\usepackage{float}
\usepackage{placeins}
\usepackage{multirow}
\usepackage[toc,page]{appendix}
\usepackage{tabularx}

\DeclareMathOperator*{\argmax}{arg\,max}

\newcommand{\states}{\mathcal{S}}
\newcommand{\actions}{\mathcal{A}}
\newcommand{\Real}{\mathbb{R}}

\title{Robust Maximum Entropy Behavior Cloning}

\author{%
  Mostafa Hussein\\
  Cognitive Assistive Robotics Lab\\
  University of New Hampshire\\
  Durham, NH 03801 \\
  \texttt{mhussein@cs.unh.edu} \\
\And
  Brendan Crowe\\
  Department of Statistics\\
  University of New Hampshire\\
  Durham, NH 03801 \\
  \texttt{bjc1041@wildcats.unh.edu} \\
\And
  Marek Petrik\\
  Department of Computer Science\\
  University of New Hampshire\\
  Durham, NH 03801 \\
  \texttt{mpetrik@cs.unh.edu} \\
\And
  Momotaz Begum\\
  Cognitive Assistive Robotics Lab\\
  University of New Hampshire\\
  Durham, NH 03801 \\
  \texttt{mbegum@cs.unh.edu} \\
}

\begin{document}

\maketitle
\vspace{-5mm}
\begin{abstract}
Imitation learning~(IL) algorithms use expert demonstrations to learn a specific task. Most of the existing approaches assume that all expert demonstrations are reliable and trustworthy, but what if there exist some adversarial demonstrations among the given data-set? This may result in poor decision-making performance. We propose a novel general frame-work to directly generate a policy from demonstrations that autonomously detect the adversarial demonstrations and exclude them from the data set. At the same time, it's sample, time-efficient, and does not require a simulator. To model such adversarial demonstration we propose a min-max problem that leverages the entropy of the model to assign weights for each demonstration. This allows us to learn the behavior using only the correct demonstrations or a mixture of correct demonstrations.
  
  
  
\end{abstract}

\vspace{-4mm}
\section{Introduction and Related Work}
\vspace{-2mm}

Imitation learning~IL addresses the problem of learning a policy from demonstrations provided by an expert \cite{chernova2014robot,osa2018algorithmic}. As robots become more involved in our daily lives, the ability to program robots and teach them new skills becomes significantly more important. The ability of a robot to effectively learn from demonstrations would greatly increase the quality of robotics applications. A common assumption in most IL approaches is that all expert demonstrations are reliable and trustworthy, but that is not always the case. In this paper we address the problem of adversarial demonstration and how we can detect those demonstrations in any given data-set. Before we go further we want to define what an adversarial demonstration is and why it might exist in a data-set. It is any demonstration that does not follow the optimal policy/policies defined by the task expert.

There are two main approaches for IL: inverse reinforcement learning~(IRL), where we learn a reward function that the demonstrator is trying to maximize during the task, then generating a policy that maximizes the generated reward \cite{ng2000algorithms,russell1998learning}. More recent approaches \cite{finn2016guided,ho2016model}, draw a connection between IRL and generative adversarial networks \cite{finn2016connection,goodfellow2014generative} and managed to get better expected return than the classical IRL algorithms. The application of these new techniques in practice is often hindered by the need for millions of samples during training to converge even in the simplest control tasks \cite{kostrikov2018discriminator}.

The second approach is behavioral cloning~(BC), the goal in BC is to learn a mapping between the states and actions as a supervised learning problem \cite{pomerleau1991efficient}. BC is considered conceptually simple and theoretically sound \cite{syed2010reduction}.  The main criticism for BC in its current state is the covariate shift \cite{ross2010efficient,ross2011reduction}. One of the main advantages of BC over IRL is, it does not require a simulator or extra samples during the learning.
To be able to deploy a robot and safely use it in our daily lives, we must have the ability to teach the robot new tasks without the need for a simulator to sample from, as well as considering the time efficiency. This feature is only feasible using BC and that is our main reason behind building our approach upon BC.

A few works like \cite{zheng2014robust} assume the existence of noisy demonstrations and propose a Bayesian approach to detect them, the authors use a latent variable to assign a weight to each data point in the demonstration set and find these weights using an EM-like algorithm. Criticism of this approach is that they use an assumption over a prior distribution which is mostly task dependent and they can only handle until 10\% of the data is random noise, and cannot handle structured adversarial behavior. Other approaches in IRL like \cite{grollman2011donut,shiarlis2016inverse} use the ``failed'' demonstration to train the model beside the correct ones, but they assume that these failed demonstrations are given and labeled in the demonstration set.

In this paper, we propose a novel robust probabilistic IL frame-work that has the ability to autonomously detect the adversarial demonstrations and exclude it from the training data-set. Robust Maximum ENTropy~(RM-ENT), is a frame-work that defines the demonstrated task by constraining the feature expectation matching between the demonstration and the generated model. The feature matching constraint by itself cannot generate a policy and here is where the maximum entropy principles \cite{amari2016information,jaynes1957information} will play the main role in our frame-work. (1) It will choose the model among the task model space that has the maximum entropy; (2) Simultaneously it will analyze the entropy contributed by each demonstration and will set weights to each demonstration that distinguishes between the correct and adversarial ones. We demonstrate that RM-ENT achieves better expected return and robustness than existing IRL and standard BC in classical control tasks in the OpenAi-gym simulator \cite{brockman2016openai}.

\vspace{-3mm}
\section{Preliminaries and Base Model}
\vspace{-3mm}

We use a tuple $(\mathcal{S},\mathcal{A},\rho_0)$ to define an infinite horizon Markov process~(MDP), where $\mathcal{S}$ represents the state space, $\mathcal{A}$ represents the action space, $\rho_0: \mathcal{S} \rightarrow \mathbb{R}$ is the distribution of the initial state $s_0$. Let $\pi$ denote a stochastic policy $\pi : \mathcal{S} \times \mathcal{A} \rightarrow [0,1]$ and $\pi_E$ denote the expert policy we have from the demonstrations. The expert demonstrations $\mathcal{D}$ are a set of trajectories, each of which consists of a sequence of state-action pairs $\mathcal{D}=(a_i,s_i)_{i=1}^Q$ where $Q$ is the number of state-action pairs in each demonstration.

In most IL algorithms we try to represent the task using a set of features $f_i(s,a), i \in \{1,2,\ldots,n \}$ that contain enough information to help us solve the IL problem while limiting the complexity of learning. Now comes the most common questions in the IL problem: \emph{What should we match between the expert and the learner?}
Many answers have been introduced among the IL community but the most successful approach until now is the \emph{feature expectation matching (FEM)} \cite{ng2,finn2016guided,syed2008apprenticeship,maxent}:

\vspace{-2mm}
\begin{equation}
\label{eqn2}
\begin{aligned}
\mathbb{E}_{\tilde{\pi}}[f_i] & = \mathbb{E}_{{\pi}}[f_i], i \in \{1,2,\ldots,n \}\\
\sum_{s \in \mathcal{S}}\sum_{a \in \mathcal{A}} \tilde{p}(s) \tilde{\pi}(a\vert s)f_i(s,a) &=  \sum_{s \in \mathcal{S}}\sum_{a \in \mathcal{A}} \tilde{p}(s) \pi(a \vert s)f_i(s,a)
\end{aligned}
\end{equation}
\vspace{-2mm}

Where $\tilde{p}$ is the state-action expert distribution while $p$ is the learned model and $\tilde{p}(s)$ is the expert distribution of $s$ in the demonstration set.

FEM by itself is an ill-defined problem that cannot generate a policy in the case of BC or a reward function in IRL, since there are many optimal policies that can explain a set of demonstrations, and many rewards that can explain an optimal policy.

We use the principles of maximum entropy \cite{jaynes1957information} to solve the ambiguity among the model space where we are looking for the model that had the maximum entropy with the constraint of FEM. 

\vspace{-3mm}
\begin{equation} \label{primal_org_model}
\begin{aligned}
\max_{\pi \in \mathbb{R}^{\mathcal{S}\times\mathcal{A}}} \quad & H(\pi) \equiv - \sum_{s \in \mathcal{S}}\sum_{a \in \mathcal{A}} \tilde{p}(s)\pi(a \vert s) \log \pi(a \vert s)\\
\textrm{s.t.} \quad & \mathbb{E}_{\tilde{\pi}}[f_i] - \mathbb{E}_{{\pi}}[f_i] = 0 \quad  i = 1, \ldots, n \\
& \sum_{a \in \mathcal{A}} \pi(a|s) -1 =0 \quad \forall \  s \in \mathcal{S} \\
\end{aligned}
\end{equation}
\vspace{-3mm}

Using a Lagrange multiplier we can solve this convex problem and get a generalized form for the policy.\footnote{A complete derivation can be found in Appendix A.} Using the previous formulation we manage to generate a policy using only a few demonstrations because it depends on the feature itself not on how many data points we have, which will be shown in the result section.

\vspace{-3mm}
\section{Robust Maximum Entropy Behavior Cloning (RM-ENT)}
\vspace{-3mm}

In the previous section, we introduced how to learn the best fit model from our set of demonstrations, but the assumption was that those demonstrations are coming from the expert without any noise or inaccurate trajectories which is not the case in real-life applications.
Our goal here is to be able to use only the set of the demonstration that can lead us to the optimal policy and exclude anything else.

Now we will introduce how we can add robustness to our model. We will add the $w$ variable which is a weight that is given to each demonstration. The goal is to give the adversarial demonstration the minimum possible weight and to give the correct demonstration a higher weight automatically through the learning. The main hypothesis is coming from maximum entropy principles. The original definition of entropy is the average level of uncertainty inherent in the random variable. So we can say that we are looking for the demonstrations that add the least amount of entropy to the model.

We can explain more by saying, if we have an adversarial demonstration it will try to add incorrect, or ``random'', information to the model which will increase its entropy. So the goal is to limit this adversarial demonstration by assigning a lower weight to it. At the same time, if two demonstrations add the same amount of information to the model, they should have the same weight. Based on the previous discussion we will introduce these two new notations:

\vspace{-7mm}
\begin{subequations}
	\begin{tabularx}{\textwidth}{Xp{0.0mm}X}
	\begin{equation}
	\tilde{p}_{w}(s) = \frac{1}{M} \displaystyle\sum_{d=1}^D w_{d} \cdot  \tilde{p}(s|d)
	\end{equation}
	& &
	\begin{equation}
	\tilde{\pi}_{w}(s|a) = \frac{1}{M}\displaystyle\sum_{d=1}^D w_{d} \cdot \tilde{\pi}(s,a|d)
	\end{equation}
	\end{tabularx}
\end{subequations}
\vspace{-5mm}

Where $D$ is the total number of demonstrations, and $M$ should be $\sum_{d=1}^D w_{d}$. Which is the minimum number of demonstrations that we can trust in the given set.

By modifying \eqref{primal_org_model} with the new variable $w$ we will get our primal problem as follows:

\vspace{-7mm}
	\begin{equation}
	\begin{aligned} 
	\min_{w \in \Real^D} \max_{\pi \in \Real^{\states\times\actions}} \quad &  - \sum_{s \in \mathcal{S}}\sum_{a \in \mathcal{A}} \pi(a \vert s) \log \pi(a \vert s) \sum_{d=1}^{D} w_{d} \cdot \tilde{p}(s,d) \\
	\operatorname{s.\,t.} \ 
	\quad & \sum_{d =1}^D  w_{d} \sum_{s \in \mathcal{S}}\sum_{a \in \mathcal{A}} f_{i}(s,a) \tilde{p} (s,d) \Big(\pi(a\vert s) - \tilde{\pi} (a\vert s,d) \Big)= 0,\quad i = 1,\ldots,N  \qquad \textcolor{gray}{[\pi]}\\
	&\sum_{a \in \mathcal{A}} \pi(a \vert s) -1 =0, \quad \forall s \in \mathcal{S} \qquad \textcolor{gray}{[\pi]} \\
	& \sum_{d=1}^D w_{d} = M , \quad w_{d} \geq 0, \quad \forall d \in \mathcal{D}, \quad w_{d} \leq 1 \quad \forall d = 1, \dots, D \qquad \textcolor{gray}{[w]}\\
	\end{aligned}
	\end{equation}
\vspace{-8mm}

Using a Lagrange multiplier we can solve this problem.\footnote{A complete derivation and more details about the optimization algorithm can be found in Appendix B.}

\vspace{-3mm}
\section {Experiments and Results}\label{Exp}
\vspace{-3mm}
\subsection{Experiments with Grid world} \label{openai-gym}

In our first experiment, we used a $5 \times 5 $ grid world as a toy example where the agent starts from the lower-left grid square and has to make its way to the upper-right grid square. In this experiment we mainly want to study the effect of using a different type of demonstrations and how successful our frame-work is at detecting any adversarial demonstrations.

A reminder that our frame-work takes only the demonstrations as an input without any more information about its correctness and generates the policy and at the same time a $w$ weight for each input demonstration. To best show how our algorithm is robust, we used three different types of demonstrations (Correct, adversarial, and random) as shown in Fig.\ref{grid_demo} .

\begin{figure}[!ht]
	\centering
	\subfigure []{\label{p_3}
		\includegraphics[width=.25\textwidth,height=1.5cm]{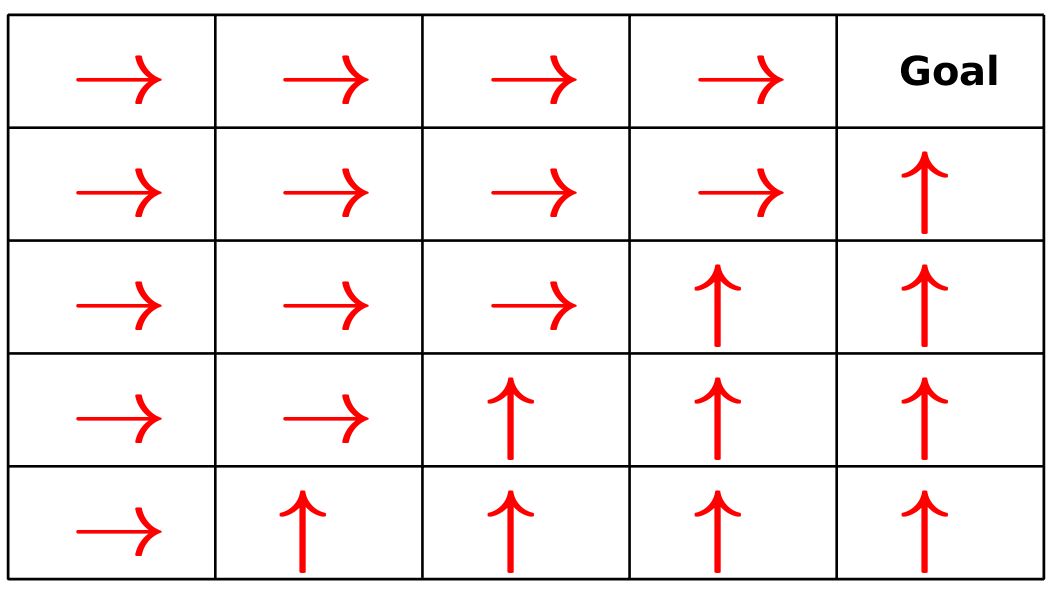}}
	\subfigure []{\label{p_4}
		\includegraphics[width=.25\textwidth,height=1.5cm]{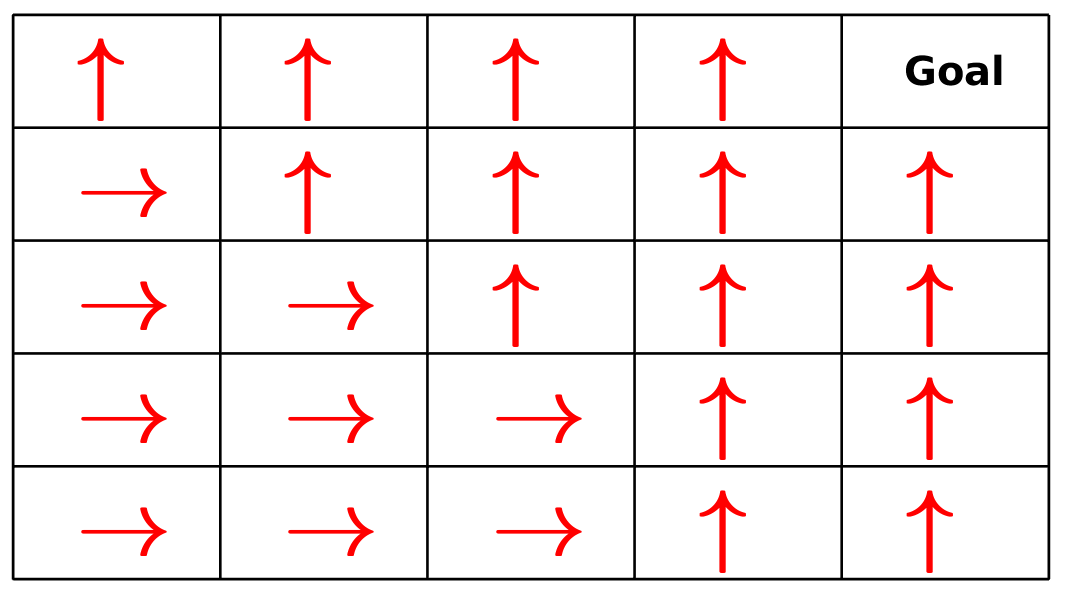}}
	\subfigure []{\label{p_5}	
		\includegraphics[width=.25\textwidth,height=1.5cm]{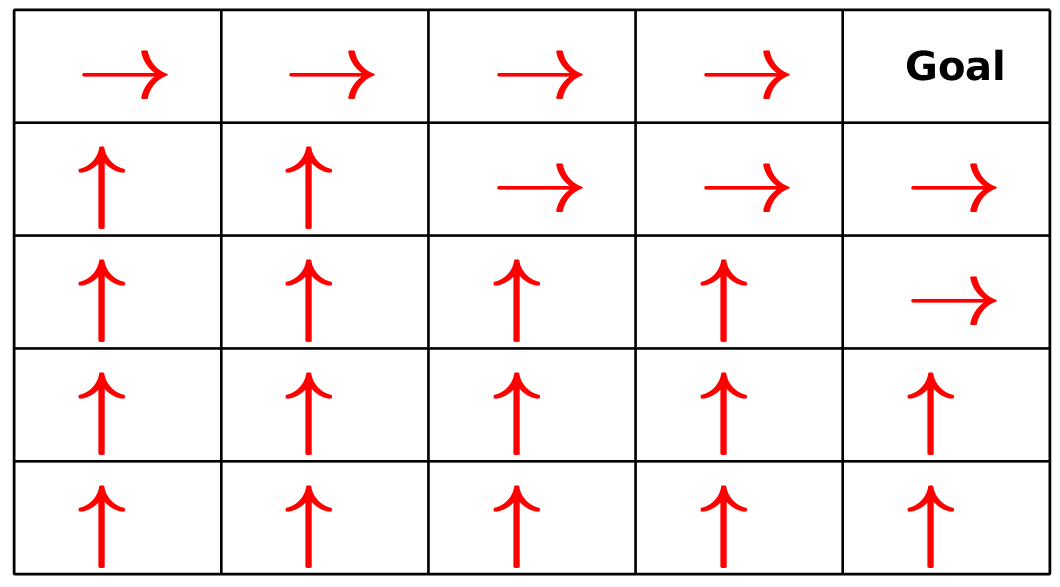}}
	\subfigure []{\label{grid_res1}	
		\includegraphics[width=.2\textwidth,height=1.7cm]{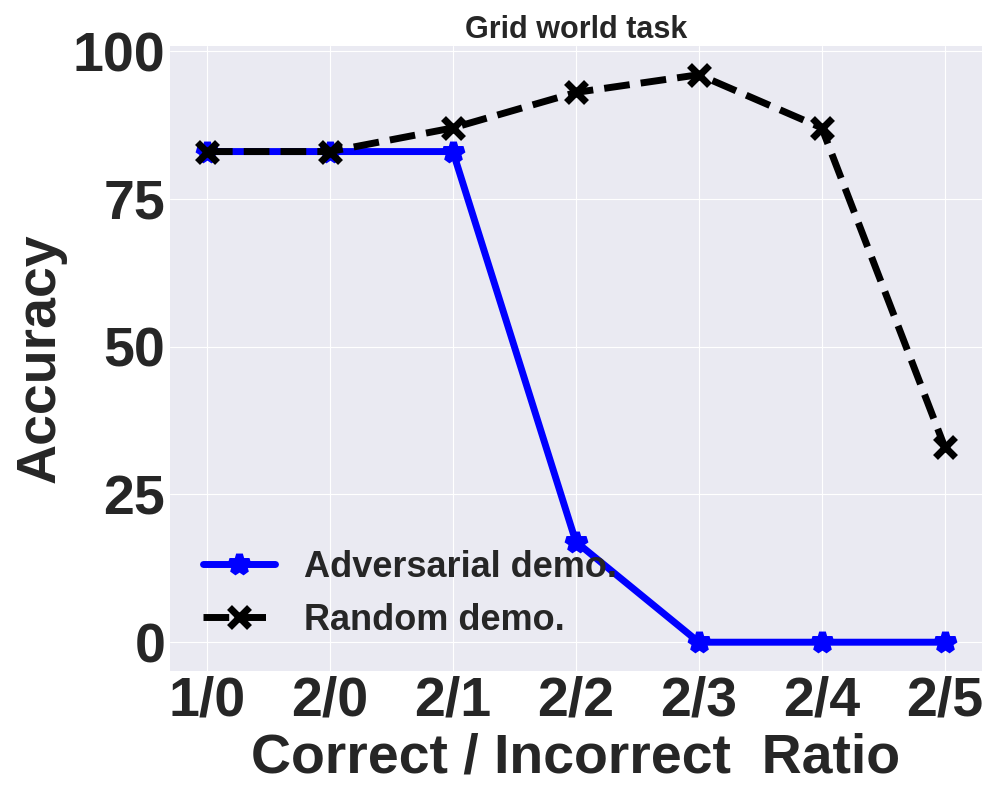}}
	
	\vspace{-3mm}	
	\caption{\ref{p_3} is the result of using both of the correct demonstration as a mixture, \ref{p_4} is the result of using correct demo. and an adversarial demo. , \ref{p_5} is the result of using correct demo. and a random demo. , \ref{grid_res1} is the accuracy using different correct/incorrect ration in case of random and adversarial demonstrations.} \label{grid_res}
\end{figure}

As shown in Table \ref{summary} \footnote{Can be found in Appendix C.} , we can see the three different cases: (1) Using two correct demonstration the algorithm correctly assigns $w=0.5$ for each demo and used both to generate the policy (accuracy = 100 \%); (2) In the second case the algorithm assigns $w=0.5$ to the two correct demonstrations and $w=0.0$ to the adversarial demonstrations(accuracy= 83 \%); (3) In the third case the algorithm assigns $w=0.5$ to the two correct demonstrations and $w=0.0$ to the random demonstrations (accuracy= 92 \%).
One last note in cases of using a random demonstrations the frame-work is able to detect those random demonstrations even if the number of correct demonstrations is less, that's because the entropy is a measurement of the randomness in the model, and the more random actions are taken the higher the entropy will be and it will be easier to detect as shown in Fig.\ref{grid_res1}.

\vspace{-4mm}
\begin{figure}[!ht]
	\centering
	\subfigure [Demo. 1 (Correct)]{\label{d_1}
	\includegraphics[width=.23\textwidth,height=1.5cm]{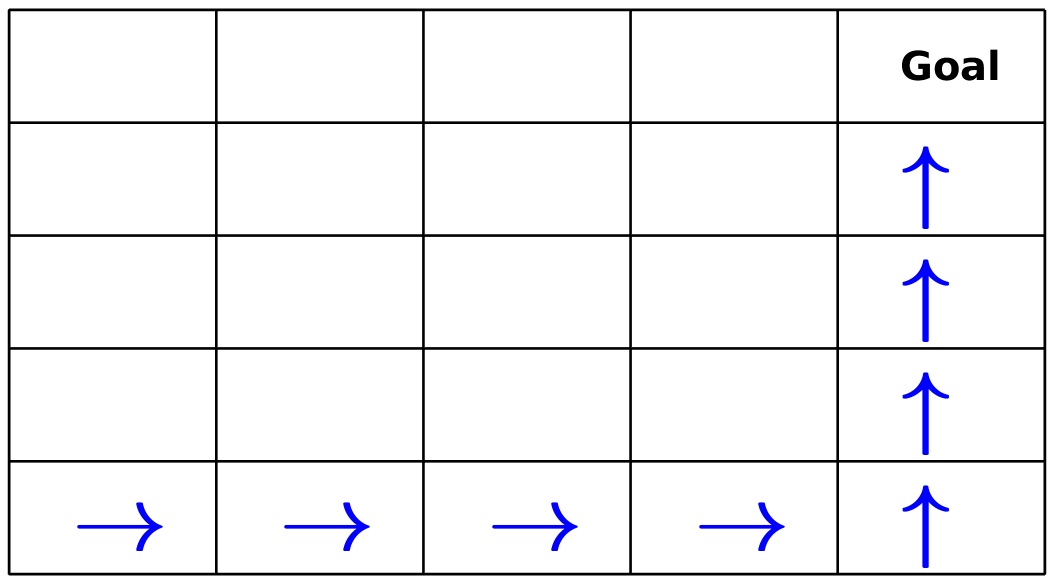}}
	\subfigure [Demo. 2 (Correct)]{\label{d_2}	
	\includegraphics[width=.23\textwidth,height=1.5cm]{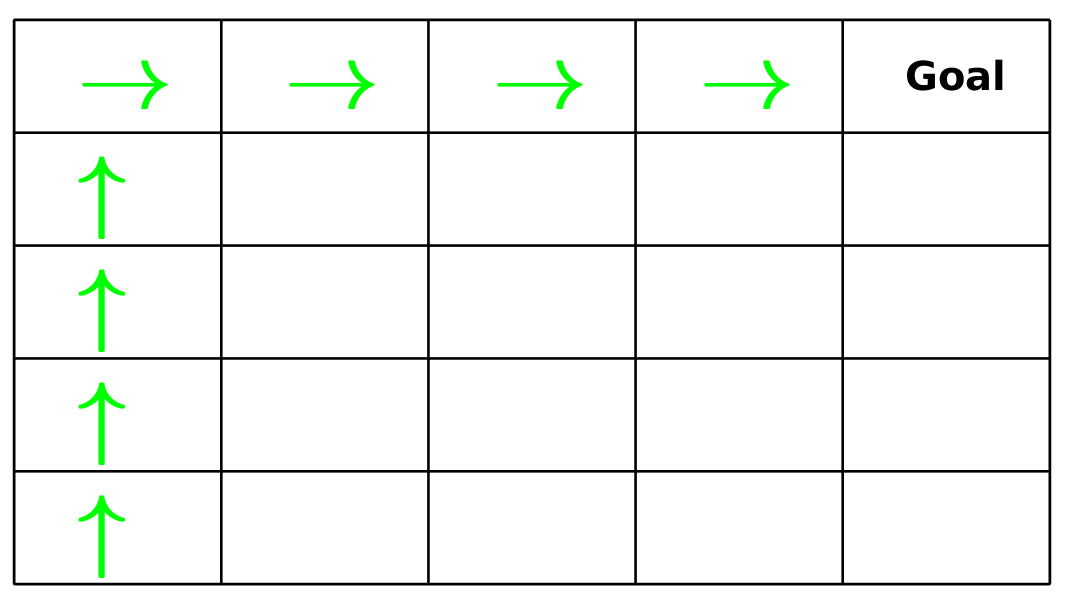}}
	\subfigure [Demo. 3 (Adversarial)]{\label{d_3}	
	\includegraphics[width=.23\textwidth,height=1.5cm]{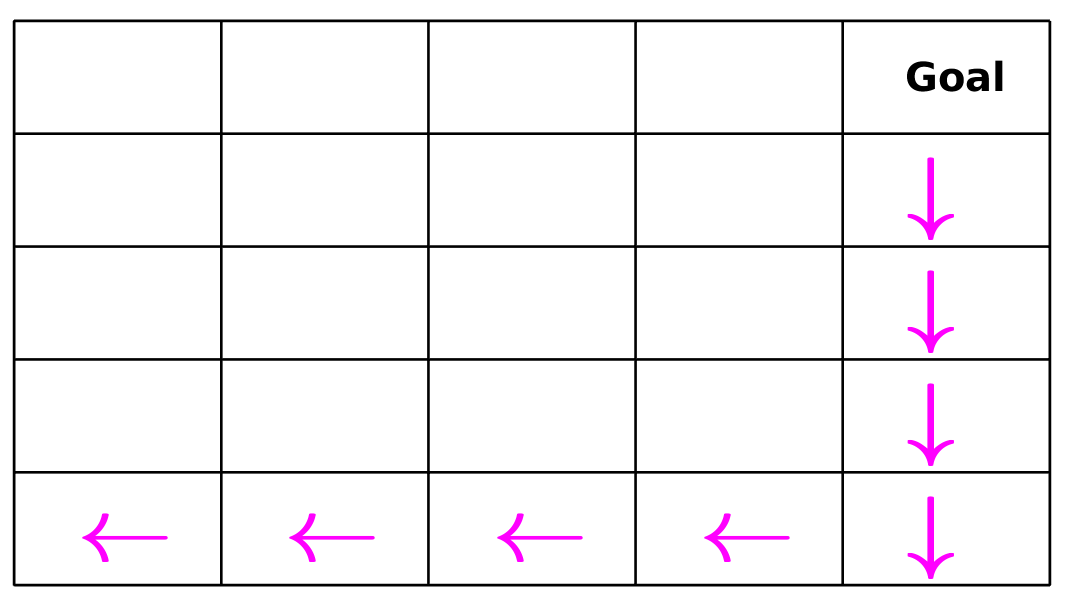}}
	\subfigure [Demo. 4 (Random)]{\label{d_4}	
	\includegraphics[width=.23\textwidth,height=1.5cm]{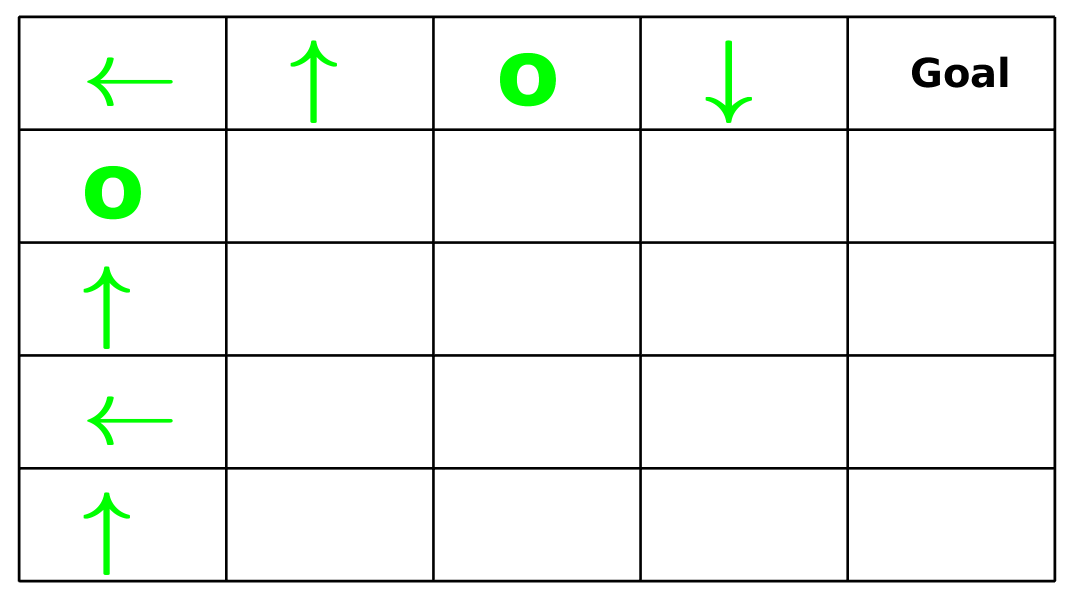}}
	
	\vspace{-4mm}	
	\caption{Demonstrations set used in the experiment.} \label{grid_demo}
\end{figure}

\vspace{-9mm}

\vspace{-5mm}
\subsection{Experiments with OpenAI-Gym Simulator} \label{openai-gym}

\vspace{-5mm}
\begin{figure}[!ht]
	\centering
	\subfigure[ ]{\label{gym_1}
		\includegraphics[scale=0.12]{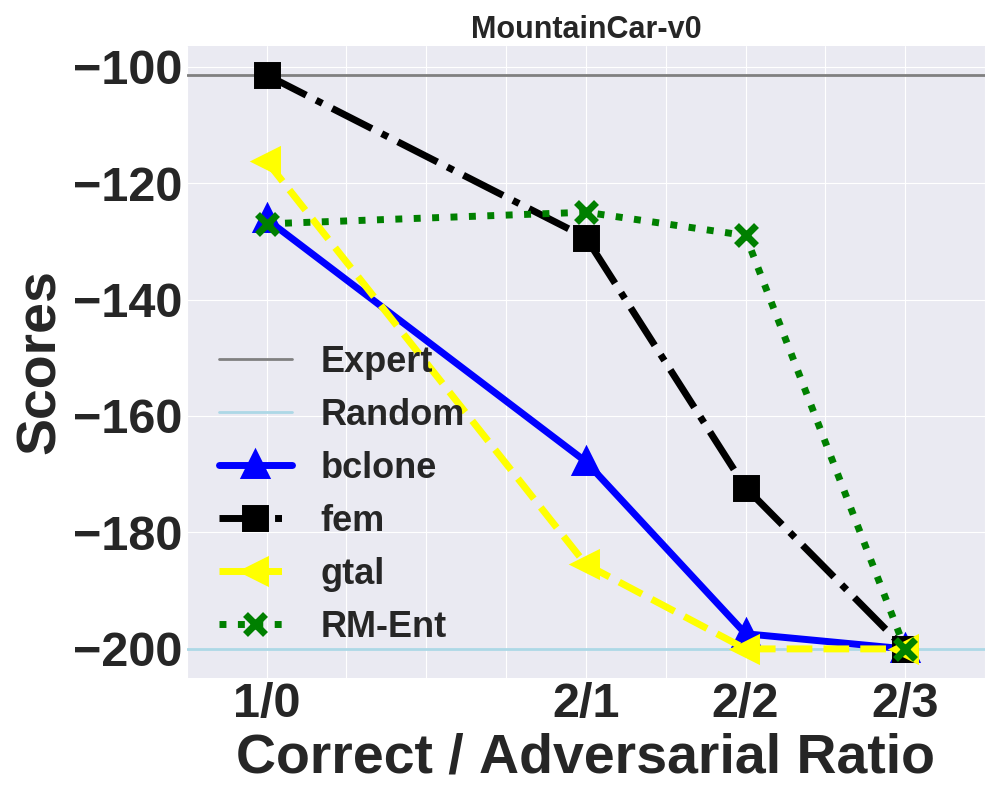}}
	\subfigure [ ]{\label{gym_2}
		\includegraphics[scale=0.12]{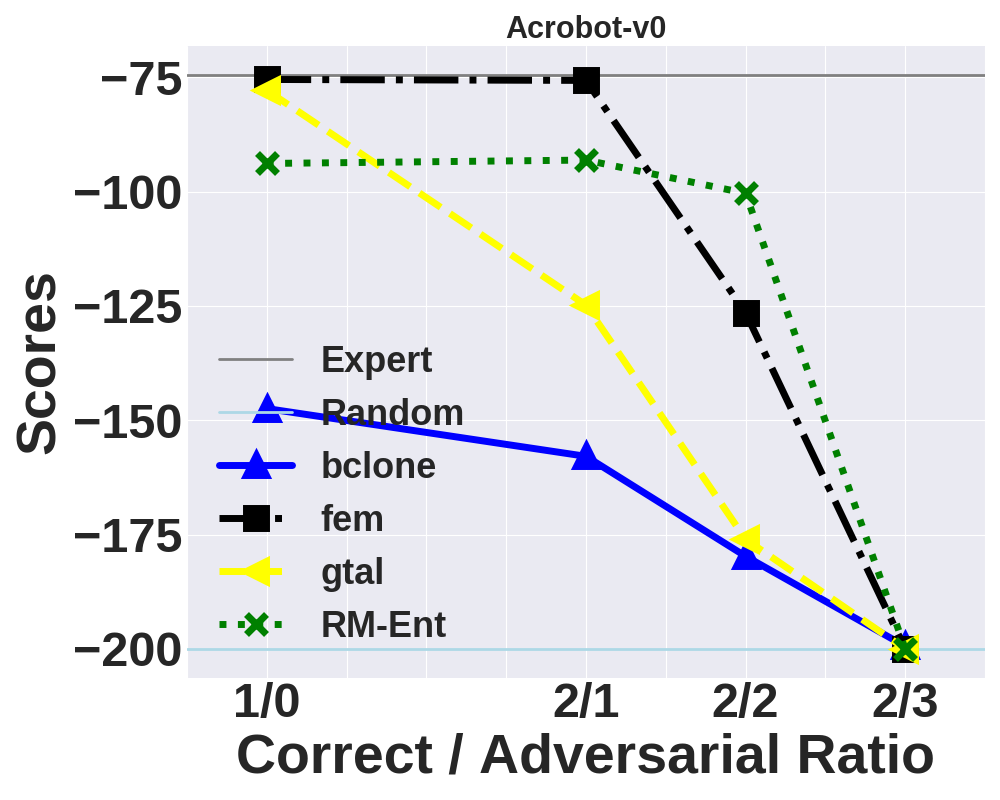}}
	\subfigure [ ]{\label{gym_time}
		\includegraphics[scale=0.12]{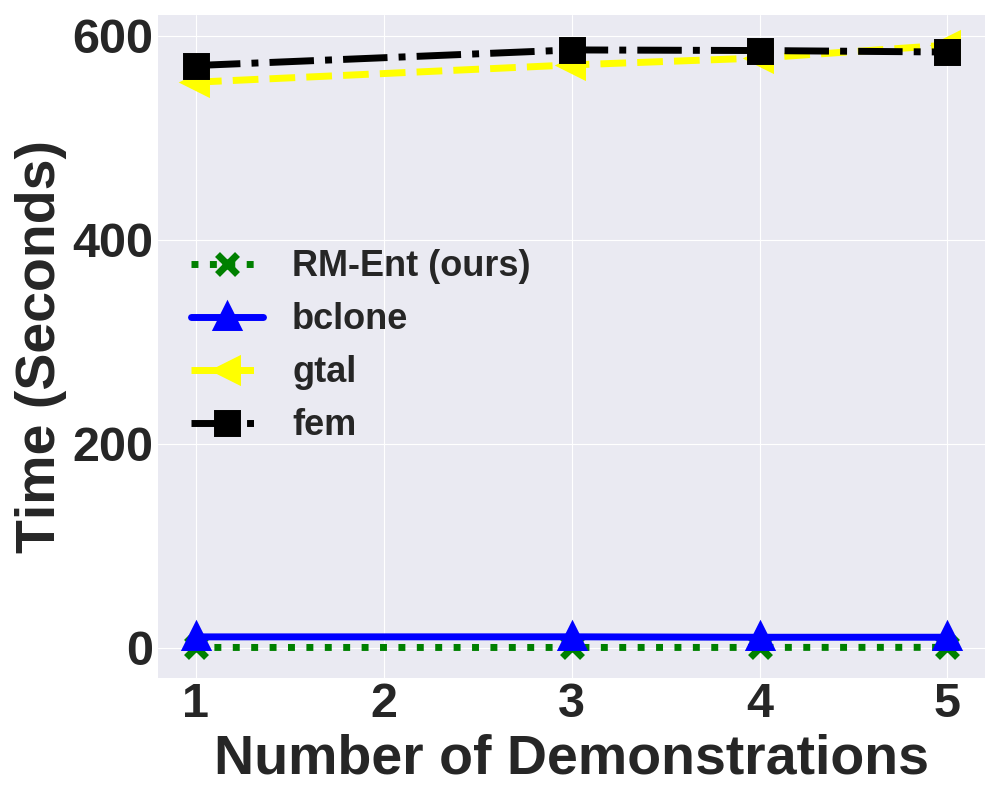}}
	\vspace{-3mm}
	\caption{Results of \textit{Mountain-Car} and \textit{Acrobot} experiments.} \label{gym_res}
\end{figure}
\vspace{-2mm}

We run our algorithm on the classical control tasks \textit{Mountain-Car} \cite{moore1990efficient} and \textit{Acrobot} \cite{geramifard2015rlpy} in the OpenAi-Gym simulator \cite{brockman2016openai}. Both tasks have a continuous state space and discrete actions. Our main opponent is \textbf{BC} \cite{bc}, we model $\pi_{BC}$ using a neural network with parameter $\theta_{BC}$ and find these parameters using maximum-likelihood estimation such that $\theta_{BC}=\argmax _{\theta}\prod_{(s,a)\in D} \pi_{BC} (a|s)$ . Also, we compared our algorithm against one of the recent approach in IRL \cite{ho2016model} with two different objective function; (1) Linear cost function from \cite{ng2}~(FEM); (2) Game-theoretic apprenticeship learning (GTAL): the algorithm of \cite{ho2016model} using the cost function from \cite{syed2008game}.\footnote{More details about the experiment parameter and number of samples can be found in Appendix C.}.

Fig. \ref{gym_1},\ref{gym_2} shows the performance of different algorithms, under varying numbers of expert and adversarial demonstrations. We can see at the first point that RM-ENT is like BC as we use only correct demonstrations. However, starting from the second point we can see the power of our algorithm as it detects that we have an adversarial demonstration among the data set and remove it (set it's weight to zero) which will keep our accuracy unchanged. While other algorithms accuracy will decrease due to the adversarial demonstration. At the final point where we have more adversarial demonstration than the correct demonstrations, all the algorithms go to a random-like policy.
We compared the time required to train each algorithm. As shown in Fig. \ref{gym_time} , RM-ENT requires much less time to converge, the reason for this is the use of neural network to train and run the opponent algorithms.

\vspace{-3mm}
\section{Conclusion and Future Work}
\vspace{-3mm}

In this work, we presented a novel frame-work that is able to automatically assign the proper weight for each of the given demonstrations and exclude the adversarial ones from the data-set. Our algorithm can achieve superior performance and sample efficiency than BC and IRL approaches in case of the presence of adversarial demonstrations. For future work, it would be enticing to use better optimization approach and  extend the frame-work to handle continuous action space.



\begin{thebibliography}{10}

\bibitem{ng2}
Pieter Abbeel and Andrew~Y Ng.
\newblock Apprenticeship learning via inverse reinforcement learning.
\newblock In {\em Proceedings of the International Conference on Machine
  Learning (ICML)}, page~1. ACM, 2004.

\bibitem{amari2016information}
Shun-ichi Amari.
\newblock {\em Information geometry and its applications}, volume 194.
\newblock Springer, 2016.

\bibitem{bc}
Michael Bain and Claude Sammut.
\newblock A framework for behavioural cloning.
\newblock In {\em Machine Intelligence 15}, pages 103--129, 1995.

\bibitem{brockman2016openai}
Greg Brockman, Vicki Cheung, Ludwig Pettersson, Jonas Schneider, John Schulman,
  Jie Tang, and Wojciech Zaremba.
\newblock Openai gym.
\newblock {\em arXiv preprint arXiv:1606.01540}, 2016.

\bibitem{chernova2014robot}
Sonia Chernova and Andrea~L Thomaz.
\newblock Robot learning from human teachers.
\newblock {\em Synthesis Lectures on Artificial Intelligence and Machine
  Learning}, 8(3):1--121, 2014.

\bibitem{finn2016connection}
Chelsea Finn, Paul Christiano, Pieter Abbeel, and Sergey Levine.
\newblock A connection between generative adversarial networks, inverse
  reinforcement learning, and energy-based models.
\newblock {\em arXiv preprint arXiv:1611.03852}, 2016.

\bibitem{finn2016guided}
Chelsea Finn, Sergey Levine, and Pieter Abbeel.
\newblock Guided cost learning: Deep inverse optimal control via policy
  optimization.
\newblock In {\em International Conference on Machine Learning}, pages 49--58,
  2016.

\bibitem{geramifard2015rlpy}
Alborz Geramifard, Christoph Dann, Robert~H Klein, William Dabney, and
  Jonathan~P How.
\newblock Rlpy: a value-function-based reinforcement learning framework for
  education and research.
\newblock 2015.

\bibitem{goodfellow2014generative}
Ian Goodfellow, Jean Pouget-Abadie, Mehdi Mirza, Bing Xu, David Warde-Farley,
  Sherjil Ozair, Aaron Courville, and Yoshua Bengio.
\newblock Generative adversarial nets.
\newblock In {\em Advances in neural information processing systems}, pages
  2672--2680, 2014.

\bibitem{grollman2011donut}
Daniel~H Grollman and Aude Billard.
\newblock Donut as i do: Learning from failed demonstrations.
\newblock In {\em 2011 IEEE International Conference on Robotics and
  Automation}, pages 3804--3809. IEEE, 2011.

\bibitem{ho2016model}
Jonathan Ho, Jayesh Gupta, and Stefano Ermon.
\newblock Model-free imitation learning with policy optimization.
\newblock In {\em International Conference on Machine Learning}, pages
  2760--2769, 2016.

\bibitem{jaynes1957information}
Edwin~T Jaynes.
\newblock Information theory and statistical mechanics.
\newblock {\em Physical review}, 106(4):620, 1957.

\bibitem{kostrikov2018discriminator}
Ilya Kostrikov, Kumar~Krishna Agrawal, Debidatta Dwibedi, Sergey Levine, and
  Jonathan Tompson.
\newblock Discriminator-actor-critic: Addressing sample inefficiency and reward
  bias in adversarial imitation learning.
\newblock {\em arXiv preprint arXiv:1809.02925}, 2018.

\bibitem{moore1990efficient}
Andrew~William Moore.
\newblock Efficient memory-based learning for robot control.
\newblock 1990.

\bibitem{ng2000algorithms}
Andrew~Y Ng, Stuart~J Russell, et~al.
\newblock Algorithms for inverse reinforcement learning.
\newblock In {\em Icml}, volume~1, page~2, 2000.

\bibitem{nocedal2006numerical}
Jorge Nocedal and Stephen Wright.
\newblock {\em Numerical optimization}.
\newblock Springer Science \& Business Media, 2006.

\bibitem{osa2018algorithmic}
Takayuki Osa, Joni Pajarinen, Gerhard Neumann, J~Andrew Bagnell, Pieter Abbeel,
  and Jan Peters.
\newblock An algorithmic perspective on imitation learning.
\newblock {\em arXiv preprint arXiv:1811.06711}, 2018.

\bibitem{pomerleau1991efficient}
Dean~A Pomerleau.
\newblock Efficient training of artificial neural networks for autonomous
  navigation.
\newblock {\em Neural computation}, 3(1):88--97, 1991.

\bibitem{ross2010efficient}
St{\'e}phane Ross and Drew Bagnell.
\newblock Efficient reductions for imitation learning.
\newblock In {\em Proceedings of the thirteenth international conference on
  artificial intelligence and statistics}, pages 661--668, 2010.

\bibitem{ross2011reduction}
St{\'e}phane Ross, Geoffrey Gordon, and Drew Bagnell.
\newblock A reduction of imitation learning and structured prediction to
  no-regret online learning.
\newblock In {\em Proceedings of the fourteenth international conference on
  artificial intelligence and statistics}, pages 627--635, 2011.

\bibitem{russell1998learning}
Stuart Russell.
\newblock Learning agents for uncertain environments.
\newblock In {\em Proceedings of the eleventh annual conference on
  Computational learning theory}, pages 101--103, 1998.

\bibitem{schulman2015trust}
John Schulman, Sergey Levine, Pieter Abbeel, Michael Jordan, and Philipp
  Moritz.
\newblock Trust region policy optimization.
\newblock In {\em International conference on machine learning}, pages
  1889--1897, 2015.

\bibitem{shiarlis2016inverse}
Kyriacos Shiarlis, Joao Messias, and SA~Whiteson.
\newblock Inverse reinforcement learning from failure.
\newblock 2016.

\bibitem{syed2008apprenticeship}
Umar Syed, Michael Bowling, and Robert~E Schapire.
\newblock Apprenticeship learning using linear programming.
\newblock In {\em Proceedings of the 25th international conference on Machine
  learning}, pages 1032--1039. ACM, 2008.

\bibitem{syed2008game}
Umar Syed and Robert~E Schapire.
\newblock A game-theoretic approach to apprenticeship learning.
\newblock In {\em Advances in neural information processing systems}, pages
  1449--1456, 2008.

\bibitem{syed2010reduction}
Umar Syed and Robert~E Schapire.
\newblock A reduction from apprenticeship learning to classification.
\newblock In {\em Advances in neural information processing systems}, pages
  2253--2261, 2010.

\bibitem{zheng2014robust}
Jiangchuan Zheng, Siyuan Liu, and Lionel~M Ni.
\newblock Robust bayesian inverse reinforcement learning with sparse behavior
  noise.
\newblock In {\em Twenty-Eighth AAAI Conference on Artificial Intelligence},
  2014.

\bibitem{maxent}
Brian~D Ziebart, Andrew~L Maas, J~Andrew Bagnell, and Anind~K Dey.
\newblock Maximum entropy inverse reinforcement learning.
\newblock In {\em Association for the Advancement of Artificial Intelligence
  (AAAI)}, volume~8, pages 1433--1438. Chicago, IL, USA, 2008.

\end{thebibliography}

\newpage

\appendix

\section{Appendix} \label{proof}

\subsection{Dual Problem Derivation}

Starting from the primal problem:
\begin{equation} \label{primal_org_model}
\begin{aligned}
\max_{\pi \in \mathbb{R}^{\mathcal{S}\times\mathcal{A}}} \quad & H(\pi) \equiv - \sum_{s \in \mathcal{S}}\sum_{a \in \mathcal{A}} \tilde{p}(s)\pi(a \vert s) \log \pi(a \vert s)\\
\textrm{s.t.} \quad & \mathbb{E}_{\tilde{\pi}}[f_i] - \mathbb{E}_{{\pi}}[f_i] = 0 \quad  i = 1, \ldots, n \\
& \sum_{a \in \mathcal{A}} \pi(a|s) -1 =0 \quad \forall \  s \in \mathcal{S} \\
\end{aligned}
\end{equation}

To derive the dual problem we will use the Lagrange method for convex optimization problems.

\begin{equation}
\Lambda(\pi,\lambda,\mu) \equiv H(\pi) + \displaystyle\sum_{i=1} ^ N \lambda_{i} \Big( \mathbb{E}_{{\pi}}[f_i]-\mathbb{E}_{\tilde{\pi}}[f_i] \Big) + \displaystyle\sum_{s \in \mathcal{S}} \tilde{p}(s) \mu_{s} \Big( \displaystyle\sum_{a \in \mathcal{A}} \pi(a|s)-1 \Big)
\end{equation}

Where $\lambda_i$, $\mu_s$ are the Lagrangian's multiplier corresponding to each constraint.

\begin{equation}
\Lambda(\pi,\lambda,\mu) \equiv - \displaystyle\sum_{s \in \mathcal{S}} \tilde{p}(s) \displaystyle\sum_{a \in \mathcal{A}}\pi(a \mid s) \log \pi(a \mid s) + \displaystyle\sum_{i=1}^N \lambda_{i} \bigg(\mathbb{E}_{{\pi}}[f_i]-\mathbb{E}_{\tilde{\pi}}[f_i] \bigg) + \displaystyle\sum_{s \in \mathcal{S}} \tilde{p}(s) \mu_{s} \Big( \displaystyle\sum_{a \in \mathcal{A}} \pi(a|s)-1 \Big)
\end{equation}

By Differentiating the Lagrangian with respect to primal variables $p(s|a)$ and letting them to be zero, we obtain:

\begin{equation}
\frac{\partial \Lambda}{\partial \pi(a|s)} = - \displaystyle\sum_{s \in \mathcal{S} } \tilde{p}(s) \Big( 1+\displaystyle\sum_{a \in \mathcal{A} } \log \pi(a|s) \Big) + \displaystyle\sum_{i=1}^N \lambda_{i} \Big( \displaystyle\sum_{s \in \mathcal{S} }\tilde{p}(s)\displaystyle\sum_{a \in \mathcal{A} }f(s,a)\Big) + \displaystyle\sum_{s \in \mathcal{S}} \tilde{p}(s) \mu_{s}
\end{equation}

\begin{equation}
- \displaystyle\sum_{s \in \mathcal{S} } \tilde{p}(s) \Big( 1+\displaystyle\sum_{a \in \mathcal{A} } \log \pi(a|s) \Big) + \displaystyle\sum_{i=1}^N \lambda_{i} \Big( \displaystyle\sum_{s \in \mathcal{S}}\tilde{p}(s)\displaystyle\sum_{a \in \mathcal{A}}f(s,a)\Big) + \displaystyle\sum_{s \in \mathcal{S}} \tilde{p}(s) \mu_{s}=0
\end{equation}

\begin{equation}
\displaystyle\sum_{s \in \mathcal{S} } \tilde{p}(s) \Bigg( -1 -\displaystyle\sum_{a \in \mathcal{A} } \log \pi(a|s) + \displaystyle\sum_{i=1}^N \lambda_{i} \Big( \displaystyle\sum_{a \in \mathcal{A}}f(s,a)\Big) + \mu_{s} \Bigg) =0
\end{equation}

Assuming $\tilde{p}(s) \neq 0$,

\begin{equation}
\log \pi(a|s) = \displaystyle\sum_{i=1}^N \lambda_{i} \Big( f_{i}(s,a) \Big) + \mu_s-1
\end{equation}

\begin{equation}
\pi(a|s) = \exp \bigg( \displaystyle\sum_{i=1}^N \lambda_{i} \Big( f_{i}(s,a) \Big) \bigg) \cdot \exp \Big( \mu_s-1 \Big)
\label{primal-var}
\end{equation}

Since $\displaystyle\sum_{a \in \mathcal{A}} \pi(a|s) = 1$

\begin{equation}
\displaystyle \sum_{a \in \mathcal{A}} \exp \bigg( \displaystyle\sum_{i=1}^N \lambda_{i} \Big( f_{i}(s,a) \Big) \bigg) \cdot \exp \Big( \mu_s-1 \Big) =1 
\end{equation}

\begin{equation}
\frac{1}{ \displaystyle\sum_{a \in \mathcal{A}} \exp \bigg( \displaystyle\sum_{i=1}^N \lambda_{i} \Big( f_{i}(s,a) \Big) \bigg) }= \exp \Big( \mu_s-1 \Big) =(z_{\lambda}(s))^{-1}
\end{equation}

By substituting in Eq \ref{primal-var} we will get. 

\begin{equation}
\pi^* (a|s) = (z_{\lambda}(s))^{-1}  \cdot \exp \bigg( \displaystyle\sum_{i=1}^N  \lambda_{i} \Big( f_{i}(s,a) \Big) \bigg)
\end{equation}

Finally, the dual problem will be:

\begin{equation} \label{max-ent-dual}
-\bigg\{ \max_{\lambda} \ \Lambda(\lambda) \equiv - \displaystyle\sum_{s \in \mathcal{S}} \tilde{p}(s) \log z_{\lambda}(s) + \displaystyle\sum_{i=1}^N \lambda_{i} \sum_{s \in \mathcal{S}}\sum_{a \in \mathcal{A}}   \tilde{\pi}(s,a)f(s,a) \bigg\}
\end{equation}

\section{Appendix }

\subsection{Dual Problem of Robust Maximum Behavior Cloning}

We will start from Eq. \ref{max-ent-dual} and build upon it.
As we mentioned in the main text we will introduce the $w$ weight as part of our model.

\begin{subequations}
	\begin{tabularx}{\textwidth}{Xp{0.0mm}X}
	\begin{equation}
	\tilde{p}_{w}(s) = \frac{1}{M} \displaystyle\sum_{d=1}^D w_{D}  \tilde{p}(s|d)
	\end{equation}

	\begin{equation}
	 \tilde{\pi}_{w}(s,a) = \frac{1}{M} \displaystyle\sum_{d=1}^D w_{D}  \tilde{\pi}(s,a|d)
	\end{equation}
	\end{tabularx}
\end{subequations}

By substituting in Eq \ref{max-ent-dual} we will get.

\begin{equation}
\begin{aligned}
\min_{w} \ \ \  -\  \bigg \{ \max_{\lambda}& \ \ \ \ \Lambda(\lambda) \equiv \frac{1}{M} \displaystyle\sum_{d=1}^N w_{d}  \Big( -\displaystyle\sum_{s \in \mathcal{S}} \tilde{\pi}_{w}(s|d) \log z_{\lambda}(s) + \displaystyle\sum_{i=1} ^N \lambda_{i} \sum_{s \in \mathcal{S}}\sum_{a \in \mathcal{A}} \tilde{\pi}_{w}(s,a|d) f(s,a)\Big) \bigg\}\\
\textrm{s.t.} \quad  & \sum_{d=1}^D w_{d} = M \\
& w_{d} \geq 0 \quad \forall d \in \mathcal{D} ={1 .... D}\\
& w_{d} \leq 1 \quad \forall d \in \mathcal{D} ={1, \ldots, D}\\
\end{aligned}
\end{equation}

For simplification let's assume:

\begin{equation}
a_{d}=\displaystyle\sum_{s \in \mathcal{S}} \tilde{\pi}(s|d) \log z_{\lambda}(s)
\end{equation}

\begin{equation}
b_{d}=\displaystyle\sum_{i=1}^N \lambda_{i} \sum_{s \in \mathcal{S}}\sum_{a \in \mathcal{A}} \tilde{\pi}(s,a|d) f(s,a)
\end{equation}

\begin{equation}
c_{d}=b_{d}-a_{d}  \quad \forall d \in D ={1, \ldots, D}
\end{equation}

\begin{equation}
\begin{aligned}
\min_{w} \ \ \  -\  \Bigl\{ \max_{\lambda}  & \Lambda(\lambda) \equiv  \frac{1}{M} \sum_{d=1}^D w_{d} c_{d} \Bigr\}\\
\textrm{s.t.} \quad & \displaystyle\sum_{d=1}^N w_{d} = M \\
& w_{d} \leq 1 \quad \forall d \in \mathcal{D} \\
& w_{d} \geq 0 \quad \forall d \in \mathcal{D}
\end{aligned}
\end{equation}

By moving the negative sign to inside we will reach our final optimization problem.

\begin{equation}
\begin{aligned}
\min_{\lambda, w} \ \ \ \ & \Lambda(\lambda) \equiv - \frac{1}{M} \displaystyle\sum_{d=1}^D w_{d} c_{d} \\
\textrm{s.t.} \quad & \displaystyle\sum_{d=1}^D w_{d} = M \\
& w_{d} \leq 1 \quad \forall d \in D ={1 .... D}\\
& w_{d} \geq 0 \quad \forall d \in D ={1 .... D}
\end{aligned}
\end{equation}

From the last equation, we can see its a non-convex problem, we used Sequential Quadratic Programming~(SQP) approach to solve this problem, the basic SQP algorithm is described in chapter 18 of Nocedal and Wright \cite{nocedal2006numerical}.

SQP approach allows you to closely mimic Newton's method for constrained optimization just as is done for unconstrained optimization. At each major iteration, an approximation is made of the Hessian of the Lagrangian function using a quasi-Newton updating method. This is then used to generate a Quadratic Programming~(QP) subproblem whose solution is used to form a search direction for a line search procedure. we leveraged the function implementation in Matlab and used it to solve our problem. \footnote{https://www.mathworks.com/help/optim/ug/constrained-nonlinear-optimization-algorithms.html\#bsgppl4}

\section{Appendix}

\subsection{Grid World Experiment}

\setlength{\arrayrulewidth}{0.1mm}
\renewcommand{\arraystretch}{1.1}

\begin{table}[!ht]
	\centering
	\caption{Results of grid world}  
	\begin{tabular}{ | c | c | c | c | c | c | }
		\hline
		\textbf{Experiment  } & \textbf{Demo. number} & \textbf{Demo. Type} & \textbf{Accuracy} & \textbf{Weights} & \textbf{Cor. / Adv.} \\ \hline
		
		\multirow{2}{*}{Mixture of correct}    & demo.\_1, Fig.\ref{d_1} & ``correct'' & \multirow{2}{*}{100 \% Fig.\ref{p_3}}    &  0.5  &  \multirow{2}{*}{2/0} \\ \cline{2-3} \cline{5-5}
		&  demo.\_2, Fig.\ref{d_2} & ``correct''  &     &  0.5  & \\ \hline                                 
		
		\multirow{3}{*}{Correct \& Adversarial} &   demo.\_1, Fig.\ref{d_1}  & ``correct'' &  \multirow{3}{*}{83 \% Fig.\ref{p_4}}   &  0.5  &  \multirow{3}{*}{2/1} \\ \cline{2-3} \cline{5-5}
		&  demo.\_2, Fig.\ref{d_1} &  ``correct''  &     &  0.5  &  \\ \cline{5-5} \cline{2-3}
		&  demo.\_3, Fig.\ref{d_3}  & ``adversarial''  &     &  0.0  &  \\ \hline	
		
		\multirow{4}{*}{Correct \& Random} & demo.\_1, Fig.\ref{d_2} & ``correct''   &   \multirow{4}{*}{92 \% Fig.\ref{p_5}}  &  0.5  &  \multirow{3}{*}{2/3} \\ \cline{2-3} \cline{5-5}
		& demo.\_2, Fig.\ref{d_2}  & ``correct''   &     &  0.5  &  \\ \cline{5-5} \cline{2-3}
		& demo.\_3, Fig.\ref{d_4}  & ``random''  &     &   0.0 &  \\ \cline{5-5}\cline{2-3}
		& demo.\_4, Fig.\ref{d_4}  & ``random''  &     &   0.0 &  \\ \cline{5-5}\cline{2-3}
		& demo.\_5, Fig.\ref{d_4} &  ``random''  &     &   0.0 &  \\ \hline
	\end{tabular}
	\label{summary}
\end{table}

\subsection{Classical Control Experiments in OpenAI-Gym Simulator Details}

The expert data was generated using TRPO \cite{schulman2015trust} on the true cost functions. For the adversarial demonstrations, we simply manipulated the actions of the expert data. For example, in the mountain car, we had two actions 0,1. If the expert data was taking action 0  with a specific observation we replaced it with action 1 and vice versa. The idea behind that is to generate an adversarial demonstration that tries to fool the algorithm.

For a fair comparison, we used the same experimental settings as in \cite{ho2016model}, including the exact neural network architectures for the policies and the optimizer parameters for TRPO \cite{schulman2015trust} for all of the algorithms except ours which do not use any neural network.

The amount of environment interaction used for FEM and GTAL is shown in Table \ref{sample} . A reminder that BC and RM-ENT do not use any more samples during the training.

\begin{table}[t]
	\centering
	\caption{Parameter for FEM and GTAL} 
	\label{sample}
	\begin{tabular}{ | c | c | c | }
		\hline
		\textbf{Task} & \textbf{Training iterations } & \textbf{State-action pairs per iteration}  \\ \hline
		Mountain Car & 300 & 5000 \\ \hline
		
		Acrobot & 300 & 5000 \\ \hline

	\end{tabular}
\end{table}

\end{document}